\documentclass[times, twoside, watermark]{zreview-styletex}
\usepackage{blindtext}
\usepackage{cuted}
\usepackage[hang,flushmargin]{footmisc} 
\pgfplotsset{compat=1.16}

\newcounter{texexp}
\tcbset{
texexp/.style={colframe=myblue!80!black, colback=myblue!5!white,
fonttitle=\small\sffamily\bfseries},
example/.code 2 args={\refstepcounter{texexp}\label{#2}%
\pgfkeysalso{texexp,title={Box \thetexexp: #1}}},
}
\newenvironment{texexptitled}[3][]{\tcblisting{example={#2}{#3},#1}}{\endtcblisting}
\usepackage[para]{threeparttable}

\leadauthor{Vidal} 

\begin{document}

\title{Perspectives on individual animal identification \\from biology and computer vision}
\shorttitle{Animal ID Review}

% Use letters for affiliations, numbers to show equal authorship (if applicable) and to indicate the corresponding author
\author[1,2]{Maxime Vidal}
\author[3]{Nathan Wolf}
\author[3]{Beth Rosenberg}
\author[3]{Bradley P. Harris}
\author[1,2]{Alexander Mathis}
\affil[1]{Center for Neuroprosthetics, Center for Intelligent Systems, Swiss Federal Institute of Technology (EPFL), Lausanne, Switzerland}
\affil[2]{Brain Mind Institute, School of Life Sciences, Swiss Federal Institute of Technology (EPFL), Lausanne, Switzerland}
\affil[3]{Fisheries, Aquatic Science, and Technology (FAST) Laboratory, Alaska Pacific University, Anchorage, AK, USA}
\affil[*]{Corresponding author: alexander.mathis@epfl.ch}

\maketitle

%TC:break Abstract
%the command above serves to have a word count for the abstract
\begin{abstract}
Identifying individual animals is crucial for many biological investigations. In response to some of the limitations of current identification methods, new automated computer vision approaches have emerged with strong performance. Here, we review current advances of computer vision identification techniques to provide both computer scientists and biologists with an overview of the available tools and discuss  their applications. We conclude by offering recommendations for starting an animal identification project, illustrate current limitations and propose how they might be addressed in the future.
\end {abstract}

\section*{Introduction}

The identification\footnote{In publications, the terminology \textit{re-identification} is often used interchangeably. In this review we posit that \textit{re-identification} refers to the recognition of (previously) known individuals, hence we use \textit{identification} as the more general term.} of specific individuals is central to addressing many questions in biology: does a sea turtle return to its natal beach to lay eggs? How does a social hierarchy form through individual interactions? What is the relationship between individual resource-use and physical development? Indeed, the need for identification in biological investigations has resulted in the development and application of a variety of identification methods, ranging from genetic methods~\cite{palsboll1999genetic, avise2012molecular}, capture-recapture~\cite{royle2013spatial, choo2020best}, to GPS tracking~\cite{baudouin2015identification} and radio-frequency identification~\cite{bonter2011applications, weissbrod2013automated}. While each of these methods is capable of providing reliable re-identification, each is also subject to limitations, such as invasive implantation or deployment procedures, high costs, or demanding logistical requirements. Image-based identification techniques using photos, camera-traps, or videos offer (potentially) low-cost and non-invasive alternatives. However, identification success rates of image-based analyses have traditionally been lower than the aforementioned alternatives. \medskip

Using computer vision to identify animals dates back to the early 1990s and has developed quickly since (see~\citet{schneider2019past} for an excellent historical account). The advancement of new machine learning tools, especially deep learning for computer vision~\cite{lecun2015deep,norouzzadeh2018automatically, schneider2019past, wu2020recent, Mathis2020APO}, offers powerful methods for improving the accuracy of image-based identification analyses. In this review, we introduce relevant background for animal identification with deep learning based on visual data, review recent developments, identify remaining challenges and discuss the consequences for biology, including ecology, ethology, neuroscience, and conservation modeling. We aimed to create a review that can act as a reference for researchers, who are new to animal identification and can also help current practitioners interested in applying novel methods to their identification work. 

\section*{Biological context for identification}

Conspecific identification is crucial for most animals to avoid conflict, establish hierarchy, and mate (e.g.,~\cite{levrero, martin, hagey}). For some species, it is understood how they identify other individuals --- for instance, penguin chicks make use of the distinct vocal signature based on frequency modulation to recognize their parents within enormous colonies~\cite{jouventin1999finding}. However, for many species the mechanisms of conspecific identification are poorly understood. What is certain, is that animals use multiple modalities to identify each other, from audition, to vision and chemosensation~\cite{levrero, martin,hagey}. Much like animals use different sensors, techniques using non-visual data have been proposed for identification.
\medskip 

\begin{figure*}
    \centering
        \includegraphics[width=\linewidth]{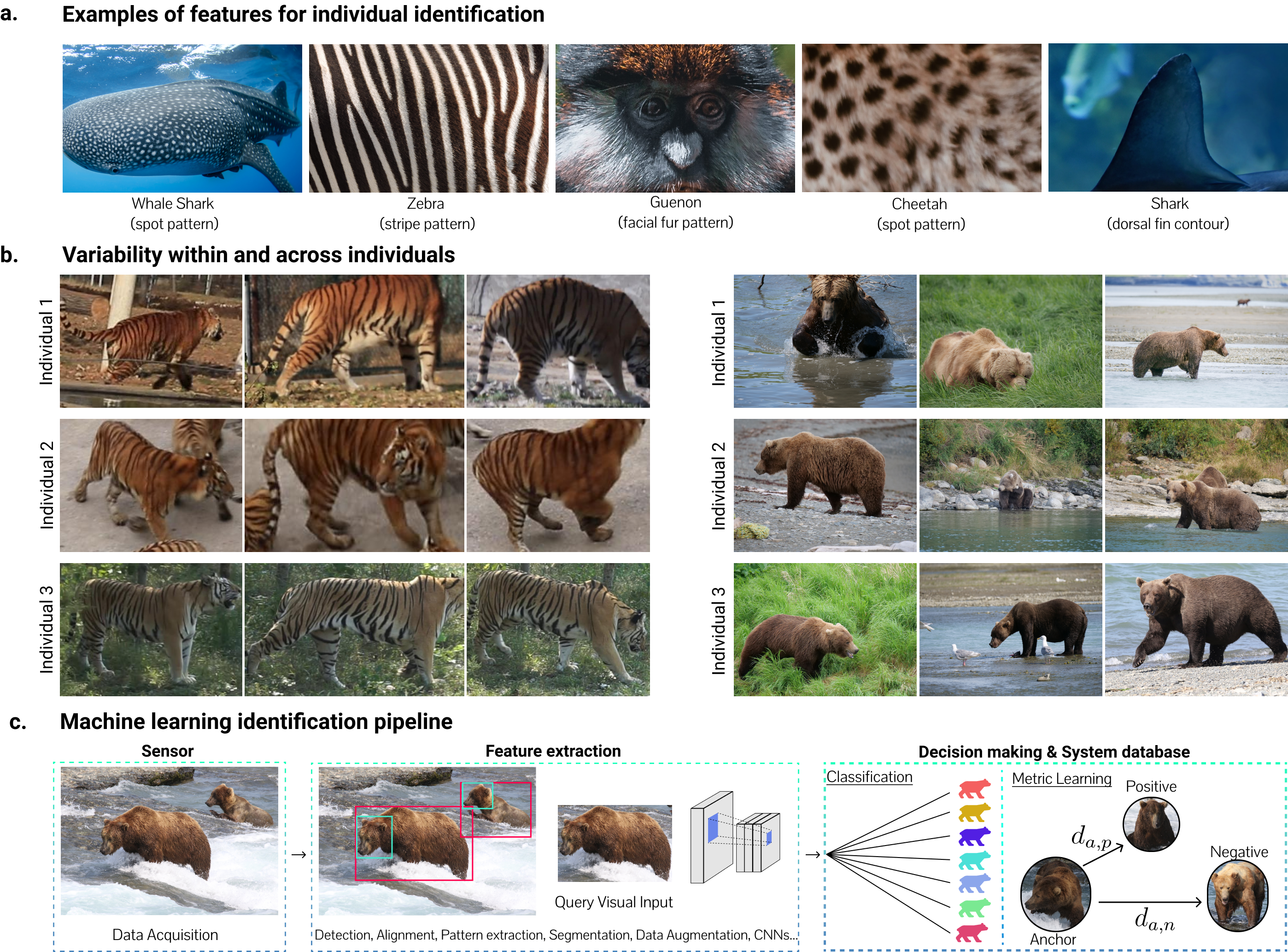}
        \caption{\textbf{a,} Animal biometrics examples featuring unique distinguishable phenotypic traits (adapted with permission from \url{unsplash.com}). \textbf{b,} Three pictures each of three example tigers from the Amur Tiger reID Dataset~\cite{li2019amur} and three pictures each of three example bears from the McNeil River State Game Sanctuary (photo credit Alaska Department of Fish and Game). The tiger stripes are robust visual biometrics. The bear images highlight the variations across seasons (fur and weight changes). Postures and contexts vary more or less depending on the species and dataset. \textbf{c,} Machine Learning identification pipeline from raw data acquisition through feature extraction to identity retrieval.}        \label{fig:pattern}
\end{figure*}
%Dataset Copyright for Amur tiger: The whole dataset is released under the non-commercial/research purposed CC BY-NC-SA 4.0 Lcense, with MakerCollider and WWF Amur tiger and leopard conservation programme team keeping the copyright of the raw video clips and all derived images

From the technical point of view, the selection of characteristics for animal identification (termed~\textit{biometrics}) is primarily based on universality, uniqueness, permanence, measurability, feasibility and reliability~\cite{jain2007handbook}. More specifically, reliable biometrics should display little intra-class variation and strong inter-class variation. Fingerprints, iris scans, and DNA analysis are some of the well-established biometric methods used to identify humans~\cite{jain2007handbook, palsboll1999genetic, avise2012molecular}. However, other physical, chemical, or behavioral features such as gait patterns may be used to identify animals based on the taxonomic focus and study design~\cite{jain2007handbook, biometrics}. For the purposes of this review, we will focus on visual biometrics and what is currently possible. 
\medskip

\section*{Visual biometrics: framing the problem}

What are the key considerations for selecting potential ``biometric'' markers in images? We believe they are: (a) a strong differentiation among individuals based on their visible traits, and (b) the reliable presence of these permanent features by the species of interest within the study area. Furthermore, one should also consider whether they will be applied to a closed or open set~\cite{phillips2011evaluation}. Consider a fully labeled dataset of unique individuals. In \textit{closed set} identification, the problem consists of images of multiple, otherwise known, individuals, who shall be ``found again'' in (novel) images. In the more general case of \textit{open set} identification, the (test) dataset may contain previously unseen individuals, thus permitting the formation of new identities. Depending on the application, both of these cases are important in biology and may require the selection of different computational methods.

\section*{Animal identification: the computer vision perspective}

\justify Some animals have specific traits, such as characteristic fur patterns, a property which greatly simplifies visual identification, while other species lack a salient, distinctive appearance (Figure~\ref{fig:pattern}a-b). Apart from visual appearance, additional challenges complicate animal identification, such as changes to the body over time, environmental changes and migration, deformable bodies, variability in illumination and view, as well as obstruction (Figure~\ref{fig:pattern}b).
\medskip

Computational pipelines for animal identification consist of a sensor and modules for feature extraction, decision-making, and a system database (Figure~\ref{fig:pattern}c;~\citep{jain2007handbook}). Sensors, typically cameras, capture images of individuals which are transformed into salient, discriminative features by the feature extraction module.  In computer vision, a feature is a distinctive attribute of the content of an image (at a particular location). Features might be e.g., edges, textures, or more abstract attributes. The decision-making module uses the computed features to identify the most similar known identities from the system database module, and in some cases, assign the individual to a new identity.
\medskip

Computer vision pipelines for many other (so-called) tasks, such as animal localization, species classification and pose estimation follow similar principles (see Box~\ref{box-tasks} for more details on those systems). As we will illustrate below, many of these tasks also play an important role in identification pipelines; for instance animal localization and alignment is a common component (see Figure~\ref{fig:pattern}c). 

 \begin{strip}
\begin{texexptitled}[text only]{Other relevant computer vision tasks}{box-tasks}
Deep learning has greatly advanced many computer vision tasks relevant to biology~\cite{lecun2015deep,norouzzadeh2018automatically, schneider2019past, wu2020recent, Mathis2020APO}. For example: \\
 \textbf{Animal detection}: A subset of \textit{object detection}, the branch of computer vision that deals with the tasks of localizing and identifying objects in images or videos. Current state of the art methods for object recognition usually employ anchor boxes, which represent the target location, size, and object class, such as in EfficientDet~\cite{efficientdet}, or newly end-to-end like, as in DETR~\cite{detr}. Of particular interest for camera-trap data is the recently released MegaDetector~\cite{beery2019efficient}, which is trained on more than 1 million labeled animal images and also actively updated\footnotemark. Relevant for camera-traps,~\citet{beery2020context} developed attention-based detectors that can reason over multiple frames, integrating contextual information and thereby strongly improving performance. Various detectors have been used in the animal identification pipeline~\cite{yolo,fasterrcnn,ssd}, which, however, are no longer state-of-the-art on detection benchmarks.\\
 \textbf{Animal species classification}: The problem of classifying \textit{species} based on pictures~\cite{speciesdlid,speciesdlid2}. As performance is correlated to the amount of training data, most recently synthetic animals have been used to improve the classification of rare species, which is a major challenge~\cite{beery2020synthetic}.\\
 \textbf{Pose estimation}: The problem of estimating the pose of an entity from images or videos. Algorithms can be top down, where the individuals are first localized, as in~\citet{hrnet} or bottom up (without prior localization) as in~\citet{higherhrnet}. Recently, several user-friendly and powerful software packages for pose estimation with deep learning of animals were developed, reviewed in~\citet{Mathis2020APO}; real-time methods for closed-loop feedback are also available~\cite{kane2020real}.\\
 \textbf{Alignment}:
In order to effectively compare similar regions and orientations - animals (in pictures) are often aligned using pose estimation or object recognition techniques.
\end{texexptitled}
\end{strip}
%\textbf{Tracking}: The problem of estimating the location of one (or multiple) entities through consecutive video frames~\cite{dell2014automated}. This can be by data association, ~\cite{romero2019idtracker}.

\footnotetext{\href{https://github.com/microsoft/CameraTraps/blob/master/megadetector.md}{https://github.com/microsoft/CameraTraps/blob/master/megadetector.md}}

In order to quantify identification performance, let us define the relevant evaluation metrics. These include top-\textit{N} accuracy, i.e., the frequency of the true identity being within the $N$ most confident predictions, and the mean average precision (mAP) defined in Box~\ref{box-glossaryDL}. A perfect system would demonstrate a top-1 score and mAP of $100\%$. However, animal identification through computer vision is a challenging problem, and as we will discuss, algorithms typically fall short of this ideal performance. Research often focuses on one species (and dataset), which is typically encouraged by the available data. Hence few benchmarks have been established, and adding to the varying difficulty of the different datasets, different evaluation methods and train-test splits are used, making the comparison between the different methods arduous. \medskip

As reviewed by~\citet{schneider2019past}, the use of computer vision for animal identification dates back to the early 1990s. This recent review also contains a comprehensive table summarizing the major milestones and publications. In the meantime the field has further accelerated and we provide a table with important animal identification datasets since its publication (Table~\ref{tab:1}). 
\medskip
 
In computer vision, features are the components of an image that are considered significant. In the context of animal identification pipelines (and computer vision more broadly), two classes of features can be distinguished. \textit{Handcrafted features} are a class of image properties that are manually selected (a process known as “feature engineering”) and then used directly for matching, or computationally utilized to train classifiers. This stands in contrast to \textit{deep features} which are automatically determined using learning algorithms to train hierarchical processing architectures based on data~\cite{lecun2015deep,wu2020recent,Mathis2020APO}. In the following sections, we will structure the review of relevant papers depending on the use of handcrafted and deep features. We also provide a glossary of relevant machine learning terms in Box~\ref{box-glossaryDL}.

\begin{table*}\centering
\ra{1.3}
\begin{threeparttable}
\begin{tabular}{@{}llllllp{0.17\linewidth}@{}}\toprule Method & Species & Target & Identities & Train Images & Test Images & Results\\\midrule
~\citet{panda} & Panda & Face & 218 & 5,845 & 402 & Top-1:$96.27$\textsuperscript{\textbf{\textasteriskcentered}},$92.12$\textsuperscript{\textbf{\textdagger}}\\
~\citet{li2019amur}& Tiger (ATRW)& Body & 92 & 1,887 & 1,762 & Top-1:$88.9$, Top-5:$96.6$, mAP:$71.0$\textsuperscript{\textbf{\ddag}}\\
~\citet{liu2019part}& Tiger (ATRW) & Body & 92 & 1,887 & 1,762 & Top-1:$95.6$, Top-5:$97.4$, mAP:$88.9$\textsuperscript{\textbf{\ddag}}\\
~\citet{manta}& Manta Ray & Underside & 120 & 1,380 & 350 & Top-1:$62.05\pm3.24$, Top-5:$93.65\pm1.83$\\
~\citet{manta}& Humpback Whale& Fluke & 633 & 2,358 & 550 & Top-1:$62.78\pm1.6$, Top-5:$93.46\pm0.63$\\
~\citet{dolphins} & Common Dolphin & Fin & 180 & {\raise.17ex\hbox{$\scriptstyle\sim$}}2,800 & {\raise.17ex\hbox{$\scriptstyle\sim$}}700 & Top-1:$90.5\pm2$, Top-5: $93.6\pm1$\\
~\citet{seal} & Saimaa Ringed Seal & Pelage & 46 & 3,000 & 2,000 & Top-1:$67.8$, Top-5:$88.6$\\
~\citet{Schofield2019scienceadv}& Chimpanzee & Face & 23 & 3,249,739 & 1,018,494 & Frame-acc:$79.12\%$, Track-acc: $92.47\%$\\
\citet{clapham2020automated} & Brown Bear & Face & 132 & 3,740 & 934 & Acc: $83.9\%$\\
\bottomrule
\end{tabular}
  \begin{tablenotes}
         \footnotesize
         \item[\textasteriskcentered] Closed Set \item[\textdagger] Open Set \item[\ddag] Single Camera Wild 
       \end{tablenotes}
     \end{threeparttable}
\caption{Recent animal identification publications and relevant data. This table extends the excellent list in~\citet{schneider2019past} by subsequent publications.}

\label{tab:1}
\end{table*}

\subsection*{Handcrafted features}

% (fur pattern-based)
\justify The use of handcrafted features is a {\it powerful, classical computer vision method}, which has been applied to many different species that display unique, salient visual patterns, such as zebras stripes~\cite{zebras}, cheetahs' spots~\cite{cheetahs}, and guenons' face marks~\cite{guenons}  (Figure~\ref{fig:pattern}a).%were used to distinguish among individuals of those species.  
~\citet{3dtiger} exploited the properties of tiger stripes to calculate similarity scores between individuals through a surface model of tigers' skins. The authors report high model performance estimates (a top-1 score of $95\%$ and a top-5 score of $100\%$ on 298 individuals). It is notable that this technique performed well despite differences in camera angle of up to 66 degrees and image collection dates of 7 years, both of which serve to illustrate the strength of this approach. 
In addition to the feature descriptors used to distinguish individuals by fur patterns, these models may also utilize edge detectors, thereby allowing individual identification of marine species by fin shape. Indeed,~\citet{sharks} employed edge detection to examine great white shark fins by encoding fin contours with boundary descriptors. The authors achieved a top-1 score of $82\%$, a top-10 score of $91\%$, and a mAP of $0.84$ on 2456 images of 85 individuals~\cite{sharks}.
Similarly,~\citet{integralcurve} used an integral curvature representation of cetacean flukes and fins to achieve a top-1 score of $95\%$ using 10,713 images of 401 bottlenose dolphins and a top-1 score of $80\%$ using 7,173 images of 3,572 humpback whales. Furthermore, work on great apes has shown that both global features (i.e., those derived from the whole image) and local features (i.e., those derived from small image patches) can be combined to increase model performance~\cite{loos1, loos2}. Local features were also used in~\citet{lemur}, who achieved top-1 scores of $93.3\% \pm 3.23\%$ on a dataset of 462 images of 80 individual red-bellied lemurs.
Prior to matching, the images were aligned with the help of manual eye markings.
\medskip

Common handcrafted features like SIFT~\cite{lowe2004distinctive}, which are designed to extract salient, invariant features from images can also be utilized. Building upon this, instead of focusing on a single species,~\citet{crall2013hotspotter} developed HotSpotter, an algorithm able to use stripes, spots and other patterns for the identification of multiple species.
\medskip

As these studies highlight, for species with highly discernible physical traits, handcrafted features have shown to be accurate but often lack robustness. Deep learning has strongly improved the capabilities for animal identification, especially for species without clear visual traits. However, as we will discuss, hybrid systems have been emerged recently that combine handcrafted features and deep learning.

\subsection*{Deep features}

\justify In the last decade, deep learning, a subset of machine learning in which decision-making is performed using learned features generated algorithmically (e.g., empirical risk minimization with labeled examples; Box~\ref{box-glossaryDL}) has emerged as a powerful tool to analyze, extract, and recognize information. This emergence is due, in large part, to increases in computing power, the availability of large-scale datasets, open-source and well-maintained deep learning packages and advances in optimization and architecture design~\cite{lecun2015deep, schneider2019past, wu2020recent}. Large datasets are ideal for deep learning, but data augmentation, transfer learning and other approaches reduce the thirst for data~\cite{lecun2015deep, schneider2019past, wu2020recent, Mathis2020APO}. Data augmentation is a way to artificially increase dataset size by applying image transformations such as cropping, translating, rotating, as well as incorporating synthetic images~\cite{lecun2015deep, Mathis2020APO, beery2020synthetic}. Since identification algorithms should be robust to those changes, augmentation often improves performance. Transfer learning is commonly used to benefit from pre-trained models (Box~\ref{box-glossaryDL}).
\medskip

Through deep learning, models can learn multiple increasingly complex representations within their progressively deeper layers, and can achieve high discriminative power. Further, as deep features do not need to be specifically engineered and are learned correspondingly for each unique dataset, deep learning provides a potential solution for many of the challenges typically faced in individual animal identification. Such challenges include species with few natural markings, inconsistencies in markings (caused by changes in pelage, scars, etc.), low-resolution sensor data, odd poses, and occlusions. Two methods have been widely used for animal identification with deep learning: \textit{classification} and \textit{metric learning}.

\subsection*{Classification models}

\justify In the classification setting, a class (identity) from a set number of classes is probabilistically assigned to the input image. This assignment decision comes after the extraction of features usually done by convolutional neural networks (ConvNets), a class of deep learning algorithms typically applied to image analyses. Note that the input to ConvNets can be the raw images, but also the processed handcrafted features. In one of the first appearances of ConvNets for individual animal classification,~\citet{freytag} improved upon work by~\citet{loos2} by increasing the accuracy with which individual chimpanzees could be identified from two datasets of cropped face images (C-Zoo and C-Tai) from $82.88\pm1.52\%$ and $64.35\pm1.39\%$ to $91.99\pm1.32\%$ and $75.66\pm0.86\%$.~\citet{freytag} used linear support vector machines (SVM) to differentiate features extracted by AlexNet~\cite{krizhevsky2012imagenet}, a popular ConvNet, without the use of aligned faces. They also tackled additional tasks including sex prediction and age estimation. Subsequent work by~\citet{gorillaid} also used AlexNet features on cropped faces of gorillas, and SVMs for classification. They reported a top-5 score of $80.3\%$ with 147 individuals and 2,500 images. A similar approach was developed for elephants by~\citet{elephantid}. The authors used the YOLO object detection network~\cite{yolo} to automatically predict bounding boxes around elephants' heads (see Box~\ref{box-tasks}). Features were then extracted with a ResNet50~\cite{resnet} ConvNet, and projected to a lower-dimensional space by principal component analysis (PCA), followed by SVM classification.
On a highly unbalanced dataset (i.e., highly uneven numbers of images per individual) consisting of 2078 images of 276 individuals,~\citet{elephantid} achieved a top-1 score of $56\%$, and a top-10 score of $80\%$. This increased to $74\%$ and $88\%$ for top-1 and top-10, respectively, when two images of the individual in question were used in the query. In practice, it is often possible to capture multiple images of an individual, for instance with camera traps, hence multi-image queries should be used when available.
\medskip

Other examples of ConvNets for classification include work by~\citet{primateface}, who explored both open- and closed-set identification for 3,000 face images of 129 lemurs, 1,450 images of 49 golden monkeys, and 5,559 images of 90 chimpanzees. The authors used manually annotated landmarks to align the faces, and introduced the PrimNet model architecture,% trained end-to-end (i.e., feature detectors and decision-maker are trained jointly) with the additive margin (AM) softmax loss function~\cite{am} 
which outperformed previous methods (e.g.,~\citet{facenet} and~\citet{lemur} that used handcrafted features). Using this method,~\citet{primateface}
achieved $93.76\pm0.90\%$, $90.36\pm0.92\%$ and $75.82 \pm 1.25\%$ accuracy for lemurs, golden monkeys, and chimpanzees, respectively for the closed-set. Finally,~\citet{panda} demonstrated a face classification method for captive pandas. After detecting the faces with Faster-RCNN~\cite{fasterrcnn}, they used a modified ResNet50~\cite{resnet} for face segmentation (binary mask output), alignment (outputs are the affine transformation parameters), and classification. They report a top-1 score of $96.27\%$ on a closed set containing 6,441 images from 218 individuals and a top-1 score of $92.12\%$ on an open set of 176 individuals. ~\citet{panda} also used the Grad-CAM method~\cite{gradcam}, which propagates the gradient information from the last convolutional layers back to the image to visualize the neural networks’ activations, to determine that the areas around the pandas’ eyes and noses had the strongest impact on the identification process.
\medskip

While the examples presented thus far have employed still images, videos have also been used for deep learning-based animal identification. 
Unlike single images, videos have the advantage that neighboring video frames often show the same individuals with slight variations in pose, view, and obstruction among others. While collecting data, one can gather more images in the same time-frame (at the cost of higher storage). For videos,~\citet{Schofield2019scienceadv} introduced a complete pipeline for the identification of chimpanzees, including face detection (with a single shot detector~\cite{ssd}), face tracking (Kanade-Lucas-Tomasi (KLT) tracker), sex and identity recognition (classification problem through modified VGG-M architectures~\cite{vggm}), and social network analysis. The video format of the data allowed the authors to maximize the number of images per individual, resulting in a dataset of 20,000 face tracks of 23 individuals. This amounts to 10,000,000 face detections, resulting in a frame-level accuracy of $79.12\%$ and a track-level accuracy of $92.47\%$. 
The authors also use a confusion matrix to inspect which individuals were identified incorrectly and reasons for this error. Perhaps unsurprisingly juveniles and (genetically) related individuals were the most difficult to separate.
In follow-up work,~\citet{Bain2019count} were able to predict identities of all individuals in a frame instead of predicting from face tracks. The authors showed that it is possible to use the activations of the last layer of a counting ConvNet (i.e., whose goal is to count the number of individuals in a frame) to find the spatial regions occupied by the chimpanzees. After cropping, the regions were fed into a fine-grained classification ConvNet. 
% is used to account for observation co-occurrence and spatial relationships among individual chimpanzees. 
This resulted in similar identification precision compared to using only the face or the body, but a higher recall.
\medskip

In laboratory settings, tracking is a common approach to identify individual animals~\cite{dell2014automated, weissbrod2013automated}. Recent tracking system, such as idtracker.ai~\cite{romero2019idtracker} and TRex~\cite{walter2021TRex}, have demonstrated the ability to track individuals in large groups of lab animals (fish, mice, etc.) by combining tracking with a ID-classifying ConvNet. 

\subsection*{(Deep) Metric learning}
%Triplet Loss Models}

\justify Most recent studies on identification have focused on deep metric learning, a technique that seeks to automatically learn how to measure similarity and distance between deep features. Deep metric learning approaches commonly employ methods such as siamese networks or triplet loss (Box~\ref{box-glossaryDL}).~\citet{schneider2020similarity} found that triplet loss always outperformed the siamese approach in a recent study considering a diverse group of five different species (humans, chimpanzees, humpback whales, fruit flies, and Siberian tigers); thereby they also tested many different ConvNets, and metric learning always gave better results. Importantly, metric learning frameworks naturally are able to handle open datasets, thereby allowing for both re-identification of a known individual and the discovery of new individuals.
\medskip

Competitions often spur progress in computer vision~\cite{wu2020recent, Mathis2020APO}. In 2019 the first, large-scale benchmark for animal identification was released (example images in Figure~\ref{fig:pattern}b); it poses two \href{https://cvwc2019.github.io/cfp.html}{identification challenges on the ATRW tiger dataset}: \textit{plain}, where images of tigers are cropped and normalized with manually curated bounding boxes and poses, and \textit{wild}, where the tigers first have to be localized an then identified~\cite{li2019amur}.

\begin{strip}
\begin{texexptitled}[text only]{Deep Learning terms glossary}{box-glossaryDL} %USE this box for technical terms! 
%\textbf{KNN}:
\textbf{Machine and deep learning}: Machine learning seeks to develop algorithms that automatically detect patterns in data. These algorithms can then be used to uncover patterns, to predict future data, or to perform other kinds of decision making under uncertainty~\cite{murphy2012machine}. Deep learning is a subset of machine learning that utilizes artificial neural networks with multiple layers as part of the algorithms. For computer vision problems, \textbf{convolutional neural networks (ConvNets)} are the de-facto standard building blocks. They consist of stacked convolutional filters with learnable weights (i.e., connections between computational elements). Convolutions bake translation invariance into the architecture and decrease the number of parameters due to weight sharing, as opposed to ordinary fully-connected neural networks~\cite{krizhevsky2012imagenet,lecun2015deep,resnet}.\\% We also briefly review two widely used machine learning techniques: \textbf{support-vector machine (SVM)} and \textbf{principal component analysis (PCA)}. 
\textbf{Support vector machines (SVM)}: A powerful classification technique, which learns a hyperplane to separate data points in feature spaces. Nonlinear SVMs also exist~\cite{pml1Book}.\\
\textbf{Principal component analysis (PCA)}: An unsupervised technique that identifies a lower dimensional linear space, such that the variance of the projected data is maximized~\cite{pml1Book}.\\
 \textbf{Classification network}: A neural network that directly predicts the class of an object from inputs (e.g., images). The outputs have a confidence score as to whether they correspond to the target. Often trained with a cross entropy loss, or other prediction error based losses~\cite{krizhevsky2012imagenet,vggm, resnet}.\\%. $L_c = -\sum_c y_c\text{log}(p_c)$ define parameters if used\\
 \textbf{Metric learning}: A branch of machine learning which consists in learning how to measure similarity and distance between data points~\cite{bellet2013survey} - common examples include siamese networks and triplet loss.
 \textbf{Siamese networks}: Two identical networks that consider a pair of inputs and classify them as similar or different, based on the distance between their embeddings. It is often trained with a contrastive loss, a distance-based loss, which pulls positive (similar) pairs together and pushes negative (different) pairs away: 
\[ \mathcal{L}\left(W, Y, \vec{X}_{1}, \vec{X}_{2}\right)= (1-Y) \frac{1}{2}\left(D_{W}\right)^{2}+(Y) \frac{1}{2}\left\{\max \left(0, m-D_{W}\right)\right\}^{2}\] where $D_W$ is any metric function parametrized by $W$, $Y$ is a binary variable that represents if $(\vec{X}_{1}, \vec{X}_{2})$ is a similar or dissimilar pair~\cite{hadsell2006dimensionality}.\\
 \textbf{Triplet loss}: As opposed to pairs in siamese networks, this loss uses triplets; it tries to bring the embedding of the anchor image closer to another image of the same class than to an image of a different class by a certain margin. 
 %The possible number of triplets grows cubically with the number of samples dataset, making it hard to optimize. 
 In its naive form \[\mathcal{L}=\max (d_{a,p}-d_{a,n}+\operatorname{margin}, 0)\] where $d_{a,p}$ ($d_{a,n}$) is the distance from the anchor image to its positive (negative) counterpart. As shown in~\citet{tripletloss}, models with this loss are difficult to train, and triplet mining (heuristics for the most useful triplets) is often used. %asily learns to separate trivial cases, but has more trouble with difficult examples, and only showing it hard samples makes it difficult to train. e.g., with d_an >> d_ap it doesn't learn
 One solution is \textit{semi-hard mining}, e.g., showing moderately difficult samples in large batches, as in~\citet{facenet}. Another more efficient solution is the \textit{batch hard} variant introduced in~\cite{tripletloss}, where one samples multiple images for a few classes, and then keeps the hardest (i.e., furthest in the feature space) positive and the hardest negative for each class to compute the loss. Mining the easy positives (very similar pairs),~\cite{xuan2020improved} has recently proven to obtain good results. \\ % These types of models usually have a ConvNet backbone
% \textbf{Evaluation metrics} measure model performance. They include \textbf{top-N accuracy}, i.e., the frequency of the actual (ground truth) identity being within the top N predictions, and the mean average precision (mAP).\\
 \textbf{Mean average precision (mAP)}: With precision defined as \text{$\frac{TP}{TP+FP}$} (TP: true positives, FP: false positives), and recall defined as \text{$\frac{TP}{TP+FN}$} (FN: false negative), the average precision is the area under the precision recall curve (see~\citet{pml1Book} for more information), and the mAP is the mean for all queries.\\
 \textbf{Transfer learning}: The process when models are initialized with features, trained on a (related) large-scale annotated dataset, and then finetuned on the target task. This is particularly advantageous when the target dataset consists of only few labeled examples~\cite{zhuang2020comprehensive, Mathis2020APO}. ImageNet is a large-scale object recognition data set~\cite{russakovsky2015imagenet} that was particularly influential for transfer learning. As we outline in the main text, many methods use ConvNets pre-trained on ImageNet such as AlexNet~\cite{krizhevsky2012imagenet}, VGG~\cite{vggm}, and ResNet~\cite{resnet}. 
% \label{box-glossaryDL}
\end{texexptitled}
\end{strip}

The authors of the benchmark also evaluated various baseline methods and showed that metric learning was better than classification. Their strongest method, was a pose part-based model, which based on the pose estimation subnetwork processes the tiger image in 7 parts to get different feature representations and then used triplet loss for the global and local representations. On the single camera wild setting, the authors reported a mAP of $71.0$, a top-1 score of $88.9\%$ and a top-5 score of $96.6\%$ - from 92 identities in 8,076 videos~\cite{li2019amur}. Fourteen teams submitted methods and the best contribution for the competition, developed a novel triple-stream framework~\cite{liu2019part}. The framework has a full image stream together with two local streams (one for the trunk and one for the limbs, which were localized based on the pose skeleton) as an additional task. However, they only required the part streams during training, which, given that pose estimation can be noisy, is particularly fitting for tiger identification in the wild. \citet{liu2019part} also increased the spatial resolution of the ResNet backbone~\cite{resnet}. Higher spatial resolution is also commonly used for other fine grained tasks such as human re-identification, segmentation~\cite{deeplab} and pose estimation~\cite{higherhrnet,Mathis2020APO}. With these modification, the authors achieved a top-1 score of $95.6\%$ for single-camera wild-ID, and a score of $91.4\%$ across cameras. \medskip
%(improving by more than $15\%$ over the baselines~\cite{liu2019part}). %https://cvwc2019.github.io/leaderboard.html

Metric learning has also been used for mantas with semi-hard triplet mining~\cite{manta}. Human-assembled photos of mantas' undersides (where they have unique spots) were fed as input to a ConvNet. %and also of humpback whales' flukes, are fed as input to a ConvNet.
Once the embeddings were created,~\citet{manta} used the $k$-nearest neighbors algorithm (k-NN) for identification. The authors achieved a top-1 score of $62.05\pm3.24\%$ and top-5 of $93.65\pm1.83\%$ using a dataset of 1730 images of 120 mantas. Replicating the method for humpback whales' flukes, the authors report a top-1 score of $62.78\pm1.6\%$ and a top-5 score of $93.46\pm0.63\%$ using 2908 images of 633 individual whales. Similarly,~\citet{dolphins} used batch hard triplet loss to achieve top-1 and top-5 scores of $90.5\pm2\%$ and $93.6\pm1\%$, respectively, on 3,544 images of 185 common dolphins. When using an additional 1200 images as distractors, the authors reported a drop of $12\%$ in the top-1 score and $2.8\%$ in the top-5 score. The authors also explore the impact of increasing the number of individuals and the number of images per individual, both leading to score increases. \citet{seal} applied metric learning to re-identify Saimaa ringed seals. After segmentation with DeepLab~\cite{deeplab} and subsequent cropping, the authors extracted pelage pattern features with a Sato tubeness filter, used as input to their network. Indeed,~\citet{bakliwal2020sloop} also showed that -- for some species -- priming ConvNets with handcrafted features produced better results than using the raw images. Instead of using k-NNs,~\citet{seal} adopt topologically aware heatmaps to identify individual seals - both the query image and the database images are split into patches whose similarity is computed, and among the most similar, topological similarity is checked through angle difference ranking. For 2,000 images of 46 seals, the authors achieved a top-1 score of $67.8\%$ and a top-5 score of $88.6\%$. Overall these recent papers highlight that recent work has combined handcrafted and deep learning approaches to boost the performance.

\section*{Applications of animal identification in field and laboratory settings\footnote{For the purposes of this review, we forgo discussion of individual identification in the context of the agricultural sciences, as circumstances differ greatly in those environments. However, we note that there is an emerging body of computer vision for the identification of livestock~\cite{qiao2020bilstm, andrew2020visual}.}}
\medskip

Here, we discuss the use of computer vision techniques for animal identification from a biological perspective and offer insights on how these techniques can be used to address broad and far-reaching biological and ecological questions. In addition, we stress that the use of semi-automated or full deep learning tools for animal identification is in its infancy and current results need to be evaluated in comparison with the logistical, financial, and potential ethical constraints of conventional tagging and genetic sampling methods.
\medskip

The specific goals for animal identification can vary greatly among studies and settings, objectives can generally be classified into two categories – applied and etiological – based on rationale, intention, and study design. 
\medskip 

Applied uses include those with the primary aims of describing, characterizing, and monitoring observed phenomena, including species distribution and abundance, animal movements and home ranges, or resource selection~\cite{harris, sharks, baird}. These studies frequently adopt a top-down perspective in which the predominant focus is on groups (e.g., populations), with individuals simply viewed as units within the group, and minimal interpretation of individual variability. As such, many of the modeling techniques employed for applied investigations, such as mark-recapture~\cite{royle2013spatial, choo2020best}, are adept at incorporating quantified uncertainty in identification. However, reliable identification of individuals in applied studies is essential to accurate enumeration and differentiation, when creating generalized models based on individual observations~\cite{marin2019acoustic}. 
\medskip

As such, misidentification can result in potential bias, and  substantial consequences for biological interpretations and conclusions. For example,~\citet{johansson}	demonstrated the potential ramifications of individual misclassification on capture-recapture derived estimates of population abundance using camera trap photos of captive snow leopards (Panthera uncia). The authors employed a manual identification method wherein human observers were asked to identify individuals in images based on pelage patterns. Results indicated that observer misclassification resulted in population abundance estimates that were inflated by up to one third.~\citet{hupman} also noted the potential for individual misidentification to result in under- or over-inflation of abundance estimates in a study exploring the use of photo-based mark-recapture for assessing population parameters of common dolphins (Delphinus sp.). The authors found that inclusion of less distinctive individuals, for which identification was more difficult, resulted in seasonal abundance estimates that were substantially different (sometimes lower and sometimes higher) than when using photos of distinctive individuals only.

\medskip

Many other questions, such as identifying the social hierarchy from passive observation, demand highly accurate identity tracking~\cite{weissbrod2013automated, Schofield2019scienceadv}. \citet{weissbrod2013automated} showed that due to the fine differences in social interactions even high identification rates of $99\%$ can have measurable effects on results (as social hierarchy requires integration over long time scales). Though the current systems are not perfect, they can already outperform experts. For instance,~\citet{Schofield2019scienceadv} demonstrated (on a test set, for the frame-level identification task) that both novices (around $20\%$) and experts (around $42\%$) are outperformed by their system that reaches $84\%$, while only taking 60ms vs. 130min and 55min, for novices and experts, respectively. 
\medskip

These studies demonstrate the need to 1) be aware of the specific implications of potential errors in individual identification to their study conclusions, and 2) choose an identification method that seeks to minimize misclassification to the extent practicable given their specific objectives and study design. While the techniques described in this review have already assisted in lowering identification error rates so as to mitigate this concern, for some applications they already reach sufficient accuracy (e.g., for conservation and management~\cite{guo, lemur, duyck2015sloop, Schofield2019scienceadv, berger2017wildbook}, neuroscience and ethology~\cite{romero2019idtracker,walter2021TRex} and public engagement in zoos~\cite{brookes2020dataset}). However, for many contexts, they have yet to reach the levels of precision associated with other applied techniques.
%the skink recovery programme~\cite{duyck2015sloop, bakliwal2020sloop},
\medskip

For comparison, genetic analyses are the highest current standard for individual identification in applied investigations. While genotyping error rates caused by allelic dropouts, null alleles, false alleles, etc. can vary between $0.2\%$ and $15\%$ per locus~\cite{wang2018estimating}; genetic analyses combine numerous loci to reach individual identification error rates of $1\%$~\cite{weller, baetscher2018microhaplotypes}. We stress that apart from accuracy many other variables should be considered, such as the relatively high logistical and financial costs associated with collecting and analyzing genetic samples, and the requirement to resample for re-identification. This results in sample sizes that are orders of magnitude smaller than many of the studies described above, with attendant decreases in explanatory/predictive power. Further, repeated invasive sampling may directly or indirectly affect animal behavior. Minimally invasive sampling (MIS) techniques using feces, hair, feathers, remote skin biopsies, etc. offer the potential to conduct genetic identification in a less intrusive and less expensive manner~\cite{carroll2018genetic}. MIS analyses are; however, vulnerable to genotyping errors associated with sample quality, with potential consequent ramifications to genotyping success rates (e.g. $87\%$, $80\%$, and $97\%$ for Fluidigm SNP type assays of wolf feces, wildcat hair, and bear hair, respectively; ~\citet{carroll2018genetic} and references therein).  These challenges, coupled with the increasing success rates and low financial and logistical costs of computer vision analyses, may effectively narrow the gap when selecting an identification technique. Further, in some scenarios the acceptable level of analytical error can be reduced without compromising the investigation of specific project goals, in which case biologists may find that current computer vision techniques are sufficiently robust to address applied biological questions in a manner that is low cost, logistically efficient, and can make use of pre-existing and archival images and video footage.  
\medskip

Unlike their applied counterparts, etiological uses of individual identification do not seek to describe and characterize observed phenomena, but rather, to understand the mechanisms driving and influencing observed phenomena. This may include questions related to behavioral interactions, social hierarchies, mate choice, competition, altruism, etc. (e.g.,~\cite{clapham, parsons, weissbrod2013automated,dell2014automated}). Etiological studies are frequently based on a bottom-up perspective, in which the focus is on individuals, or the roles of individuals within groups, and interpretations of individual variability often play predominant roles~\cite{diaz}. As such, etiological investigations may seek to identify individuals in order to derive relationships among individuals, interpret outcomes of interactions between known individuals, assess and understand individuals’ roles in interactions or within groups, or characterize individual behavioral traits~\cite{krasnova,constantine,kelly,Schofield2019scienceadv}. These studies are commonly done in laboratory settings, which presents some study limitations. The ability to record data and assign it to an individual in the wild may be crucial to understand the origin and development of personality~\cite{dall2012evolutionary, stamps2010development}. Characterizing behavioral variability of individuals is of great importance for understanding behavior~\cite{roche2016demystifying}. This has been highlighted in a meta-analysis that showed that a third of behavioral variation among individuals could be attributed to individual differences~\cite{bell2009repeatability}. The impact of repeatably measuring observations for single individuals can also be illustrated in the context of brain mapping. Repeated sampling of human individuals with fMRI is revealing fine-grained features of functional organization, which were previously unseen due to variability across the population~\cite{braga2017parallel}. Overall, longitudinal monitoring of single individuals with powerful techniques such as omics~\cite{chen2012personal} and brain imaging~\cite{poldrack2021diving} is heralding an exciting age for biology. 

\section*{Starting an animal identification project}

For biological practitioners seeking to make sense of the possibilities offered by computer vision, the importance of inter-disciplinary collaborations with computer scientists cannot be overstated. Since the advent of high definition camera traps, some scientists find they have hours of opportunistically collected footage, without a direct line of inquiry motivating the data collection. Collaboration with computer scientists can help to ensure the most productive analytical approach to using this footage to derive biological insights. Further, by instituting collaborations early in the study design process, computer scientists can assist biologists in implementing image collection protocols that are specifically designed for use with deep learning analyses. 
\medskip

General considerations for starting an image-based animal identification project, such as which features to focus on, are nicely reviewed by~\citet{biometrics}. Although handcrafted features can be suited for certain species (e.g., zebras), deep learning has proven to be a more robust and general framework for image-based animal identification.
However, at least a few thousand images with ideally multiple examples of each individual are needed, constituting the biggest limitation to obtaining good results. As such, data collection is a crucial part of the process. Discussion between biologists and computer scientists is fundamental and should be engaged before data collection. As previously mentioned, camera traps~\cite{camtrap, choo2020best} can be used to collect data on a large spatial scale with little human involvement and less impact on animal behavior. Images from camera traps can be used both for model training and monitored for inference. The ability of camera traps to record multiple photos/videos of an individual allows multiple streams of data to be combined to enhance the identification process (as for localization~\cite{beery2020context}). Further, camera traps minimize the potential influence of humans on animal behavior as seen in~\citet{schneider2019past}.
\medskip

Following image collection, researchers should employ tools to automatically sieve through the data, to localize animals in pictures. 
Recent powerful detection models~\citet{beery2019efficient, beery2020context}, trained on large-scale datasets of annotated images, are becoming available and generalize reasonably well to other datasets (Box~\ref{box-tasks}). Those or other object detection models can be used out-of-the-box or finetuned to create bounding boxes around faces or bodies~\cite{yolo,fasterrcnn,ssd}, which can then be aligned by using pose estimation models~\cite{Mathis2020APO}. Additionally, animal segmentation for background removal/identification can be beneficial.\medskip

Most methods require an annotated dataset, which means that one needs to label the identity of different animals on example frames; unsupervised methods are also possible (e.g.,~\citep{crall2013hotspotter, clusteringfaces}). To start animal identification, a baseline model using triplet loss should be tried, which can be improved with different data augmentation schemes, combined with a classification loss and/or expanded into more multi-task models. If attempting the classification approach, assigning classes to previously unseen individuals is not straightforward. Most works usually add a node for "unknown individual". The evaluation pipeline to monitor the model's performance has to be carefully designed to account for the way in which it will be used in practice. Of particular importance is how to split the dataset between training and testing subsets to avoid data leakage. \medskip

Also, how well a network trained with photos from professional DSLR cameras can generalize to images with largely different quality, e.g., camera traps, must be determined. In our experience this is typically not ideal, which is why it is important to get results from different cameras during training, if generalization is important. Ideally, one trains the model with the type of data that is used during deployment. However, there are also computational methods to deal with this. For human reidentification,~\citet{camtransfer} used CycleGAN to transfer images from one camera style to another, although camera traps are perhaps too different. The generalization to other (similar) species is also a path to explore.\medskip 

Other aspects to consider are the efficiency of models, even if identification is usually in an offline setting. Also, adding a ``human-in-the-loop'' approach, if the model does not perform perfectly, can still save time relative to a fully manual approach. For other considerations necessary to build a production ready system, readers are encouraged to look at~\citet{duyck2015sloop}, who created Sloop, with subsequent deep learning integration by~\citet{bakliwal2020sloop}, used for the identification of multiple species. Furthermore,~\citet{berger2017wildbook} implemented different algorithms such as HotSpotter~\cite{crall2013hotspotter} in the \href{https://wildme.org/}{Wild Me} platform, which is actively used to identify a variety of species.

\section*{Beyond image-based identification}

As humans are highly visual creatures, it is intuitive that we gravitate to image-based identification techniques. Indeed, this preference may offer few drawbacks for applied uses of individual identification in which the researcher’s perspective is the primary lens through which discrimination and identification will occur. However, the interpretive objectives of etiological uses of identification add an additional layer of complexity that may not always favor a visually based method. When seeking to provide inference on the mechanisms shaping individual interactions, etiological applications must both 1) satisfy the researcher’s need to correctly identify known individuals, and 2) attempt to interpret interactions based on an understanding of the sensory method by which the individuals in question identify and re-identify conspecifics~\cite{tibbets2007, thom, tibbetts2002}.
\medskip

Different species employ numerous mechanisms to engage in conspecific identification (e.g., olfactory, auditory, chemosensory~\cite{levrero, martin, hagey}). For example, previous studies have noted that giant pandas use olfaction for mate selection and assessment of competitors~\cite{swaisgood2004chemical, hagey}. Conversely,~\citet{schneider2018Drosophila} showed that Drosophila, which were previously assumed not to be strongly visually based, were able to engage in successful visual identification of conspecifics. Thus, etiological applications that seek to find mechanisms of animal identification must consider both the perspectives of the researcher and the individuals under study (much like Uexküll's concept of \textit{Umwelt}~\cite{uexkullstroll}), and researchers must embrace their roles as both observers and translators attempting to reconcile potential differences between human and animal perspectives. 
\medskip

Just how animals identify each other with different senses, future methods could also focus on other forms of data. Indeed, deep learning is not just revolutionizing computer vision, but problems as diverse as finding novel antibiotics~\cite{stokes2020deep} and protein folding~\cite{service2020game}. Thus, we believe that deep learning will also strongly impact identification techniques for non-visual data and make those techniques both logistically feasible and sufficiently non-invasive so as to limit disturbances to natural behaviors. Previous studies have employed techniques that are promising. For example, acoustic signals were used by ~\citet{marin2019acoustic} for counting of rock ptarmigan, and by~\citet{stowell2019automatic} in an identification method which seems to generalize to multiple bird species. Furthermore, ~\citet{kulahci} used deep learning to describe individual identification using olfactory-auditory matching in lemurs. However, this research was conducted on captive animals and further work is required to allow for application of these techniques in wild settings. 

\section*{Conclusions and outlook}

\justify Recent advances in computational techniques, such as deep-learning, have enhanced the proficiency of animal identification methods. Further, end-to-end pipelines have been created, which allow for the reliable identification of specific individuals, with, in some cases, better than human-level performance. As most methods follow a supervised learning approach, the expansion of datasets is crucial for the development of new models, as is collaboration between computer science and biological teams in order to understand the applicable questions to both fields. Hopefully, this review has elucidated the fact that lines of inquiry to one group might have previously been unknown to the other, and that interdisciplinary collaboration offers a path for future methodological developments that are analytically nimble and powerful, but also applicable, dependable, and practicable to addressing real-world phenomena. 
\medskip

As we have illustrated, recent advances have contributed to the deployment of some methods, but many challenges remain. For instance, individual identification of unmarked, featureless animals such as brown bears or primates has not yet been achieved for hundreds of individuals in the wild. Likewise, discrimination of close siblings remains a challenging computer vision individual identification problem. How can the performance of animal individual identification methods be further improved? 
\medskip

Since considerably more attention and effort has been devoted to the computer vision question of human identification, versus animal identification, this vast literature can be used as a source of inspiration for improving animal individual identification techniques. Many human identification studies experiment with additional losses in a multi-task setting. For instance, whereas triplet loss maximizes inter-class distance, the center loss minimizes intra-class distance, and can be used in combination with the former to pull samples of the same class closer together~\cite{centerloss}. Further, human identification studies demonstrate the use of spatio-temporal information to discard impossible matches~\cite{spatialtemporal}. This idea could be used if an animal has just been identified somewhere and cannot possibly be at another distant location (using camera traps' timestamps and GPS); this concept is also employed in occupancy modeling. Re-ranking the predictions has also been employed to improve performance in human-based studies using metric learning~\cite{reranking}. This approach aggregates the losses with an additional re-ranking based distance. Appropriate augmentation techniques can also boost performance~\cite{randomerasing}. In order to overcome occlusions, one can randomly erase rectangles of random pixels and random size from images in the training data set.
\medskip

Applications involving human face recognition have also contributed significantly to the development of identification technologies. Human face datasets typically contain orders of magnitude more data (thousands of identities and many more images - e.g., the YouTube Faces dataset~\cite{youtube}) than those available for other animals. One of the first applications of deep learning to human face recognition was DeepFace, which used a classification approach~\cite{deepface}. This was followed by Deep Face Recognition, which implemented a triplet loss bootstrapped from a classification network~\cite{deeprecognition} and FaceNet by~\citet{facenet} which used triplet loss with semi hard mining on large batches. FaceNet achieved a top-1 score of $95.12\%$ when applied to the Youtube Faces dataset. Some methods also showed promise for unlabeled datasets;~\citet{clusteringfaces} proposed an unsupervised method to cluster \textit{millions} of faces with approximate rank order metric. We note that this research also raises ethical concerns~\cite{van2020ethical}. Finally, benchmarks are important for advancing research and fortunately they are emerging for animal identification~\cite{li2019amur}, but more are needed. 
\medskip

Overall, broad areas for future efforts may include 1) improving the robustness of models to include other sensory modalities (consistent with conspecific identification inquiry) or movement patterns, 2) combining advanced image-based identification techniques with methods and technologies already commonly used in biological studies and surveys (e.g., remote sensing, population genetics, etc.), and 3) creating larger benchmarks and datasets, for instance, via Citizen Science programs (e.g., \href{https://emammal.si.edu}{eMammal}; \href{https://www.inaturalist.org}{iNaturalist}, \href{http://www.greatgrevysrally.com/}{Great Grevy’s Rally}). While these areas offer strong potential to foster analytical and computational advances, we caution that future advancements should not be dominated by technical innovation, but rather, technical development should proceed in parallel with, or be driven by, the application of novel and meaningful biological questions. Following a question-based approach will assist in ensuring the applicability and utility of new technologies to biological investigations and potentially mitigate against the use of identification techniques in suboptimal settings.

\subsection*{Acknowledgements}

\justify The authors wish to thank the McNeil River State Game Sanctuary, Alaska Department of Fish and Game, for providing inspiration for this review. Support for MV, BR, NW, and BPH was provided by Alaska Education Tax Credit funds contributed by the At-Sea Processors Association and the Groundfish Forum. We are grateful to Lucas Stoffl, Mackenzie Mathis, Niccol\`{o} Stefanini, Alessandro Marin Vargas, Axel Bisi, Sébastien Hausmann, Travis DeWolf, Jessy Lauer, Matthieu Le Cauchois, Jean-Michel Mongeau, Michael Reichert, Lorian Schweikert, Alexander Davis, Jess Kanwal, Rod Braga, and Wes Larson for comments on earlier versions of this manuscript.

\section*{Bibliography}
%\bibliography{zreview}

\end{document}